\documentclass[runningheads]{llncs}
\usepackage{makecell}
\usepackage[T1]{fontenc}
\usepackage{booktabs}
\usepackage{multicol}
\usepackage{mathtools}
\usepackage{amsmath}
\usepackage{amssymb}
\usepackage{multirow}
\usepackage{algorithm}
\usepackage{algorithmic}
\usepackage{graphicx}
\usepackage{xcolor}
\begin{document}
\title{Contrastive Knowledge Amalgamation for Unsupervised Image Classification}
%
\author{Shangde Gao\inst{1,2}\thanks{Contributed equally.} \and Yichao Fu \inst{1}$^\star$\and
Ke Liu\inst{1}$^\star$ \and
Yuqiang Han\inst{1,2}}
\authorrunning{Shangde Gao, et al.}
\institute{College of Computer Science and Technology, Zhejiang University, Hangzhou City, Zhejiang Province, China \and  ZJU-Hangzhou Global Scientific and Technological Innovation Center, Hangzhou City, Zhejiang Province, China \email{gaosde@zju.edu.cn,fuyichao@zju.edu.cn,lk2017@zju.edu.cn,hyq2015@zju.edu.cn}}
\maketitle
\begin{abstract}
Knowledge amalgamation (KA) aims to learn a compact student model to handle the joint objective from multiple teacher models that are  are specialized for their own tasks respectively.
Current methods focus on coarsely aligning teachers and students in the common representation space, making it difficult for the student to learn the proper decision boundaries from a set of heterogeneous teachers. Besides, the KL divergence in previous works only minimizes the probability distribution difference between teachers and the student, ignoring the intrinsic characteristics of teachers. 
    Therefore, we propose a novel Contrastive Knowledge Amalgamation (CKA) framework, which introduces contrastive losses and an alignment loss to achieve intra-class cohesion and inter-class separation.
Contrastive losses intra- and inter- models are designed to widen the distance between representations of different classes. The alignment loss is introduced to minimize the sample-level distribution differences of teacher-student models in the common representation space.
Furthermore, the student learns heterogeneous unsupervised classification tasks through soft targets efficiently and flexibly in the task-level amalgamation. 
Extensive experiments on benchmarks demonstrate the generalization capability of CKA in the amalgamation of specific task as well as multiple tasks. Comprehensive ablation studies provide a further insight into our CKA.
\end{abstract}

\section{Introduction}

Reusing pre-trained models to get lite ones for reducing computation costs of training a new one from scratch has been a trending research topic in recent years \cite{jiang2022empirical,jiang2023ptmtorrent}. Knowledge Distillation (KD) methods \cite{HintonVD15} train a light-weight target model (the \emph{``student"} model) by learning from a well-trained cumbersome model (the \emph{``teacher"} model), which improves the performance of students with any architectures compared to the models trained from scratch. 
Knowledge Amalgamation (KA) \cite{shen2019amalgamating,LuoWFHTS19} aims to train a versatile student model by transferring knowledge from multiple pre-trained teachers. The above method requires mapping teachers and student to a common representation space. The student learn similar intermediate features through the aggregated cues from the pre-trained teachers. Further, by integrating probability knowledge from pre-trained teachers using KL divergence, student can predict the joint of teachers' label sets.

However, complex optimization designs are required for heterogeneous teachers in previous works. 
Besides, direct application of previous KA methods to downstream tasks causes severe performance degradation because of domain shifts, additional noise, as well as information loss in feature projections. Moreover, due to the imperfection of pre-trained teachers and absence of human annotation, the supervision signals for students are confused. 

In this work, we endeavor to explore an efficient and effective KA scheme for unsupervised image classification. 
We aim to transfer knowledge as much as possible from pre-trained teachers who specialize in heterogeneous unsupervised image classification tasks to a compact and versatile student. 
For example, if one teacher classifies cars and the other classifies airplanes, the student should be able to classify both cars and airplanes. 
To achieve this, we first extend the contrastive learning paradigm to the knowledge fusion environment, for two reasons.
Firstly, CL can effectively push positive sample pairs together and pull negative sample pairs apart without the need for manual annotations. 
Additionally, different teacher and student models are natural augmentation schemes, and their combination significantly increases the number of positive and negative samples for training the student. 
Secondly, supervised contrastive loss models have been shown to outperform traditional cross-entropy losses\cite{KhoslaYaoJayadevaprakashFeiFei_FGVC2011}. Thus, they can be effectively used in teacher pre-training to alleviate the incompleteness and unreliability of supervising teacher models.

We propose a novel \textbf{C}ontrastive \textbf{K}nowledge \textbf{A}malgamation, refered to as CKA, by implementing the CKA framework via DNNs for unsupervised classification. 
Concretely, we first construct a common representation space based on the shared multilayer perceptron (MLP), and design contrastive and alignment losses to achieve intra-class cohesion and inter-class separation of samples. As a way of unsupervised learning, the contrastive loss intra- and inter- models aims to enlarge the distance between feature representations of different sample categories and reduce the distance between feature representations of the same sample category. 
Besides, alignment losses are proposed to minimize the sample-level distribution difference between different models. 
Apart from learning the teachers' features, a soft target distillation loss finally is designed to effectively and flexibly transfer probability knowledge from pre-trained teachers to a student, enabling the student to make inferences similar to or the same as the teachers' during task-level amalgamation.

The contributions of this work are summarized as follows:
\begin{itemize}
    \item We propose a novel model reuse paradigm to supervise the student model without annotations, named CKA, which introduces contrastive losses and an alignment loss to achieve intra-class cohesion and inter-class separation. 
    \item We design a soft target distillation loss to effectively transfer knowledge from the pre-trained teachers to a student in the output probability space.
    \item Extensive experiments on standard benchmarks demonstrate that CKA provides more accurate supervision, and is generalizable for amalgamating heterogeneous teachers.
\end{itemize}
\section{Related Works}
\subsection{Knowledge Distillation \& Knowledge Amalgamation}
Knowledge distillation (KD) \cite{HintonVD15,Zhao2022DecoupledKD} is a method of transferring knowledge from one model to another. 
However, existing approaches are still performed under a single teacher-student relationship with a sharing task, are not applicable to multiple and heterogeneous teachers. 
Knowledge amalgamation (KA) aims to acquire a compact student model capable of handling the comprehensive joint objective of multiple teacher models, each specialized in its own task.
There are two kinds of approaches: (1) \emph{Homogeneous} KA, where all teachers and students have identical network architectures \cite{shen2019amalgamating}. (2) \emph{Heterogeneous} KA, where each teacher has different architecture and specializes in its own class set \cite{VongkulbhisalVS19,LuoWFHTS19}. Among these, \cite{VongkulbhisalVS19} matches the outputs of students to the corresponding teachers, while \cite{LuoWFHTS19} aligns the features of students and teachers in a shared latent space by minimizing the maximum mean discrepancy. However, when facing with the imperfect teachers with unreliable supervisions, previous studies suffer from conflicting supervisions in the student training process, which significantly harms the performance of the student model. 
To the best of our knowledge, it is the first time to explore the CKA paradigm for unsupervised classification tasks.
\subsection{Contrastive Learning}
Contrastive Learning is an unsupervised learning method where supervision is automatically generated from the data. Currently, contrastive learning (CL) has achieved state-of-the-art performance in representation learning \cite{GrillSATRBDPGAP20,ChenK0H20,ChenH21}. SimCLR \cite{ChenK0H20} proposes the proposal by performing data augmentation on the raw input data and mapping it to a feature space, constructing a contrastive loss (i.e., InfoNCE loss) to maximize the similarity between positive pairs and minimize the similarity between negative pairs. BYOL \cite{GrillSATRBDPGAP20} and SimSiam \cite{ChenH21} extend the work by designing their losses to measure the similarity between positive samples, effectively eliminating the need for negative samples. However, all of these approaches are tailored for single-model and single-task. 
Our method extends the concept of CL to a knowledge amalgamation environment, designing intra-and inter- model contrastive losses to explore the model-agnostic semantic similarity and further apply them to downstream unsupervised multi-classification tasks.
\section{Problem Formulation}
We define the problem of knowledge amalgamation as follows. Assume that we are given $\mathcal{T}=\left\{\mathcal{T}_{t}\right\}_{t=1}^{N}$ well pre-trained teachers, where each teacher $\mathcal{T}_{t}$ specializes a distinct classification task, i.e., a set of full labeled classes $\mathcal{T}_t=(\mathcal{D}_t; \mathcal{Y}^t)$. Our proposal is to learn a versatile student with an unlabeled dataset $\mathcal{D}= \bigcup_{t=1}^
N{\mathcal{D}_{t}}$,  which is able to perform predictions over the comprehensive class set of distinct-task teachers, $\mathcal{Y}= \bigcup_{t=1}^
N{\mathcal{Y}^{t}}$. In our KA setting, $N$ tasks $\mathcal{T}=\left\{\mathcal{T}_{t}\right\}_{t=1}^{N}$ can be built for either the same or cross dataset. Without loss of generality, we assume that for any two tasks $\mathcal{T}_i, \mathcal{T}_j \in \mathcal{T}$, their specialties are totally disjoint, i.e., $\mathcal{Y}^i \cap \mathcal{Y}^j = \oslash$.
\section{Approach}
This work is aimed to build a contrastive knowledge amalgamation framework, and implement it by DNNs for unsupervised image classification. Knowledge amalgamation is particularly challenging when teacher-student structures are heterogeneous and data annotation is not available. 
\label{approach}
\begin{figure*}[t]
\centering
\includegraphics[width=0.8\textwidth]{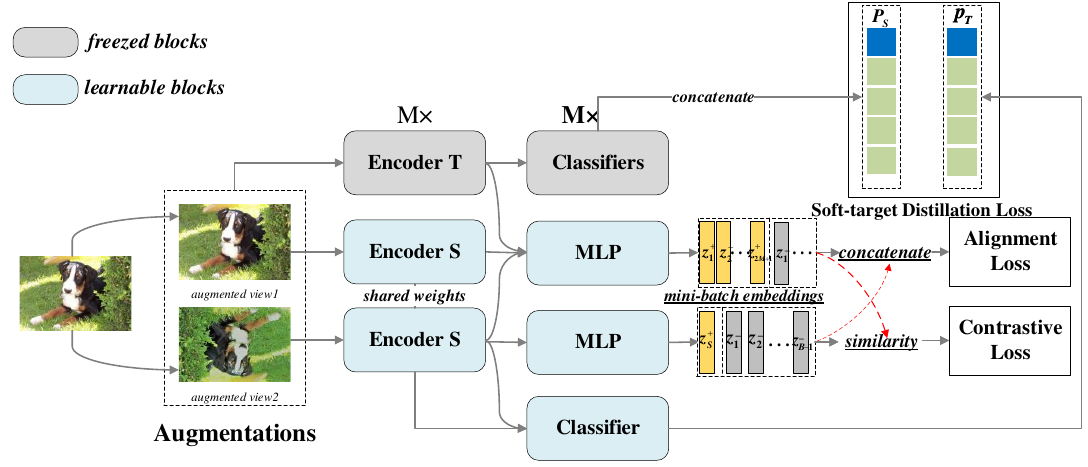} 
\caption{The overflow of contrastive knowledge amalgamation.}
\label{cka_framework}
\vspace*{-0.2in}
\end{figure*}
To tackle the difficulty, we first leverage the distance between feature representations of the samples, and introduce contrastive and alignment losses to achieve intra-class coherence and inter-class separation of the feature representations. 
Additionally, we design a soft-target distillation loss to effectively transfer the soft-target probability knowledge from pre-trained teachers to the student. The overview  of the proposed CKA is shown in Figure. \ref{cka_framework}, in which the knowledge of pre-trained teachers is fixed. By training the student model in downstream tasks, the student is capable of making inferences that are similar or identical to those of their teachers. 
\subsection{Margin-based Intra- and Inter-model Contrast}
\begin{figure*}[t]
\centering
\includegraphics[width=0.6\textwidth]{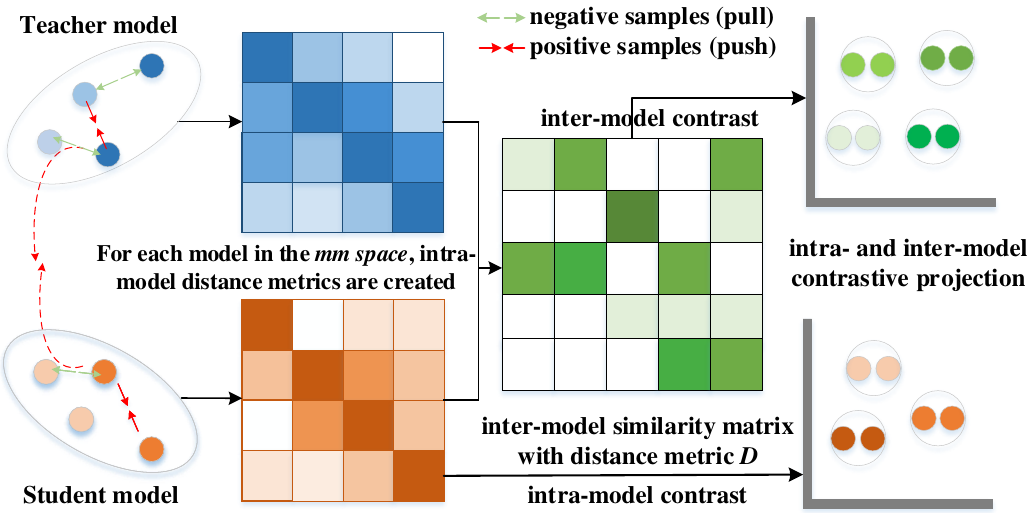}
\caption{Illustrations of intra- and inter-model contrast loss via the teacher-student pair.}
\label{model_level_contrast}
\vspace*{-0.2in}
\end{figure*}

As there are no annotated data available, we novely use contrastive learning (CL) to construct supervision for guiding the student. CL aims to maximize the similarities of positive pairs while minimizing those of negative ones \cite{ChenK0H20}. The characteristics of pairs can be defined by different criteria. Motivated by this, we develop two types of contrastive losses, including edge-based student-internal contrast (intra-model contrast) and distance-based teacher-student model contrast (inter-models contrast), to increase the distance between different sample class feature representations and decrease the distance between the same sample class feature representations. The overall schematic is shown in Figure. \ref{model_level_contrast}. 
\subsubsection{Margin-based Intra-model Contrast}
To begin with, we describe the standard contrastive loss term, following the most popular setups of SimCLR \cite{ChenK0H20}, which is defined as:
\begin{equation}
\label{base_contrast_loss}
\mathcal{L}(\tilde{x}, \hat{x}) = -\log \frac{e^{{s}(\tilde{z}, \hat{z}) / \tau}}{e^{{s}(\tilde{z}, \hat{z}) / \tau+\sum_{\bar{z} \in \Lambda^{-}} e^{{s}\left(\tilde{z}, \bar{z})\right) / \tau}}}
\end{equation}
Here, by way of randomized data augmentation Aug($\cdot$), two different views $\tilde{x}$ and $\hat{x}$ for the input sample $x$ are generated. The two images are then fed into an encoder network $\mathcal{E}(x)$, followed by a two-layer nonlinear projection head MLP $h(\cdot)$, yielding a pair of $L_{2}$-normalized positive embeddings $\hat{z} = h(\mathcal{E}(\hat{x}))$ and $\tilde{z} = h(\mathcal{E}(\tilde{x}))$. $\bar{z} \in \Lambda^{-}$ represents the negative sample in a mini-batch. $s(\cdot,\cdot)$ declares the \emph{cosine similarity} for measuring the relationship between embedding pair $\tilde{z}$ and $\hat{z}$ (resp. $\tilde{x}$ and $\hat{x}$), formulated as:
\begin{equation}
s\left(\tilde{z}, \hat{z}\right)=\frac{\left(\tilde{z}\right)\left(\hat{z}\right)^{\top}}{\left\|\tilde{z}\right\| \cdot \left\|\hat{z}\right\|}
\end{equation}
To prevent the loss from being dominated by easy negatives (different class samples with little similarity), a constant margin $\alpha$ is introduced that only negative pairs with similarity larger than $\alpha$ contribute to the contrastive loss in Eqn. \ref{base_contrast_loss}. Formally, the margin-based intra-model contrastive loss for training the student model is denoted as:
\begin{equation}
\label{margin_contrast}
\mathcal{L}_{intra} = \left( 1 - s\left(\tilde{z}, \hat{z}\right)\right) + \sum_{\bar{z} \in \Lambda^{-}}\left(s\left(\tilde{z}, \bar{z}\right) - \alpha \right)
\end{equation}
\subsubsection{Distance-based inter-model contrast}
For inter-model contrast, data across models are embedded as point distributions in high-dimensional vector spaces. 
To measure the inter-model distance between those two point distributions, we model two \emph{metric measure spaces} (mm-spaces) $\mathcal{X}=\left(X, d_{X}, \mu\right) \text { and } \mathcal{Y}=\left(Y, d_{Y}, \nu\right)$, where data $X$ (resp. $Y$) is a complete separable set endowed with a distance $d_{X}$ and a positive Borel measure $\mu \in \mathcal{M}+(X)$. 
Those two mm-spaces are considered up to isometry (denoted $\mathcal{X} \sim \mathcal{Y}$), meaning that there is a bijection $\psi: \operatorname{spt}(\mu) \rightarrow \operatorname{spt}(\nu)$ (where $\operatorname{spt}(\mu)$ is the support of $\mu$) such that $d_{X}(x, y)=d_{Y}(\psi(x), \psi(y))$ and $\psi_{\sharp} \mu=\nu$. Here $\psi_{\sharp}$ is the push-forward operator. 

Specifically, let $\mu \in \mathcal{P}\left(\mathbb{R}^{p}\right)$ and $\nu \in \mathcal{P}\left(\mathbb{R}^{q}\right)$ with $p \neq q$ to be discrete measures on mm-spaces with $\mu=\sum_{i=1}^{n} a_{i} \delta_{x_{i}}$ (here $\delta_{x_{i}}$ is the mass at $x_{i}$) and $\nu=\sum_{i=1}^{n} b_{j} \delta_{y_{j}}$ of supports $X$ and $Y$, where $a \in \Sigma_{n}$ and $b \in \Sigma_{m}$ are simplex histograms. 
The distance $\mathcal{D}$ between those points is defined as:
\begin{equation}
\footnotesize
\label{eq:distance}
\small
\mathcal{D}\left(\mathcal{X},\mathcal{Y}\right)^q=
\sum_{i, j, k, l}\left|d_{X}\left(x_{i}, x_{k}\right)-d_{Y}\left(y_{j}, y_{l}\right)\right|^{q} \pi_{i, j} \pi_{k, l}
\end{equation}

Here $d_{X}(x_i,x_k): \mathbb{R}^{p} \times \mathbb{R}^{p} \rightarrow \mathbb{R}_{+}$, 
measures the euclidean distance between sample points $x_i$ and $x_k$ in $\mu$. 
The intuition underpinning the definition of this distance is that there exists a fuzzy correspondence map $\pi \in \mathcal{P}(X \times Y)$ between the points of the distributions, which tends to associate pairs of points with similar distances within each pair: the more similar $d_{X}(x_i, x_k)$ is to $d_{Y}(y_j, y_l)$, the stronger the transport coefficients $\pi_{i,j}$ and $\pi_{k,l}$ are. From a semantic perspective, by simultaneously learning the model structures of both the teacher and student, this distance can measure the similarity between samples, reducing the distance between the feature representations of similar sample classes and increasing the distance between feature representations of dissimilar sample classes.

Given a mini-batch size $B$ of feature maps $P \in \mathbb{R}^{B \times c\times h \times w}$ and $Q_{t} \in \mathbb{R}^{B \times c\times h \times w}$ extracted from the student encoder and $t$-th teacher encoder, where $c$, $h$, and $w$ denote the number of channel, height and width of the feature maps respectively. 
For simplicity, we omit the superscripts and subscripts and denote the feature maps of two different models as $P$ and $Q$. The distance metric on $P$ and $Q$ is designed firstly to guide the contrast across different models, i.e., inter-model contrast. 
To this end, we first reshape $P$ and $Q$ to $\mathbb{R}^{B \times m}$, i.e., $P=\left[p_{1}, p_{2}, \ldots, p_{B}\right]$ and $Q=\left[q_{1}, q_{2}, \ldots, q_{B}\right]$, where $m=c\times h \times w$ is the feature vectors. 
The transport map $\pi^{p},\pi^{q}  \in \mathbb{R}^{B \times B}$  for $P$ and $Q$ can be derived by:
\begin{equation}
\label{smentic_relevance}
\pi_{i,j} = \frac{e^{-d(p_{i},p_{j})}}{\sum_{j=1}^{N}e^{-d(p_{i},p_{j})}}
\end{equation}
where $d(\cdot,\cdot)$ is the mm-space distance between two instances $p_{i}$ and $p_{j}$. 
Unless stated otherwise, euclidean distance is used in our experiments. 

As for any $k$-th row vector in $\pi^{p}$ and $\pi^{q}$, $\pi^{p}_{k}$ and $\pi^{q}_{k}$ can be termed as positive pairs because they both semantically illustrate the distance of $k$-th sample and others in the mini-batch $N$, regardless of the model representation. Our distance-based inter-model contrastive loss, discovering fine-gained sample similarity matching between the student and each teacher, can be defined as:
\begin{equation}
\label{inter_contrast_loss}
\mathcal{L}_{inter}=
\sum_{t=1}^{N}\left(\left(1 - s\left(\pi^{p}, \pi^{q}_{t\text{+}}\right)\right) +\sum_{\pi^{q}_{t\text{-}} \in \Lambda^{-}}s\left(\pi^{p}, \pi^{q}_{t\text{-}}\right)\right)
\end{equation}
where $\pi^{q}_{t\text{-}}$ and $\pi^{q}_{t\text{+}}$ denote the distance-based negative and positive pairs.
 \subsection{Common Feature Alignment} 
To enable a student to mimic the aggregated hints from heterogeneous teachers, a shared multilayer perceptron (MLP) is designed for mapping all features to a common latent space. Specifically, a $1 \times 1$ kernel convolution is added after the backbone network of each model separately, thereby unifying the outputs of different models into the same channel, which is taken to be the input of MLP and set to 256 in our implementation.

As represented in CFL \cite{LuoWFHTS19}, we adopt the Maximum Mean Discrepancy (MMD) to measure the discrepancy between the output features of the student and that of teachers in the unit ball of a reproducing kernel Hilbert space \cite{Gretton_2012}. Take a teacher-student pair as an example, we extract the mini-batch common space features with the designed shared MLP and represent them as $f_\mathcal{S}, f_\mathcal{T}\in \mathbb{R}^{B \times d}$, of which $d$ denotes the output dimension of the MLP and is set to 128 in our implementation. An empirical $l_2$ \emph{norm} approximation to the MMD distance of $f_\mathcal{S}$ and $f_\mathcal{T}$ is computed as follow:
\begin{equation}
\mathrm{MMD}= \frac{1}{B}\left\| \sum_{i=1}^{B} \phi\left(f_\mathcal{T}^{i}\right)- \sum_{j=1}^{B} \phi\left(f_\mathcal{S}^{j}\right)\right\|_{2}^{2}
\end{equation}
where $\phi$ is an explicit mapping function. The extension of multi-kernel formulation of 
MMD can then be defined as: 
\begin{equation}
\begin{aligned}
\mathrm{MMD}^{2}[K, f_\mathcal{S}, f_\mathcal{T}]= &K\left( f_\mathcal{S}, f_\mathcal{S}\right) - 2K\left( f_\mathcal{T}^i, f_\mathcal{S}^j\right)  + \\
&K\left( f_\mathcal{T}, f_\mathcal{T}\right)
\end{aligned}
\end{equation}
$K$ is defined as the convex combination of $m$ PSD kernel:
\begin{equation}
\mathcal{K}=\left\{K=\sum_{u=1}^{m} \sigma_{u} K_{u}: \sum_{u=1}^{m} \sigma_{u}=1, \sigma_{u} \geq 0, \forall u\right\}
\end{equation}
here $\mathcal{K}$ denotes the multi-prototypical kernel set. The  constraints on coefficients $\left\{\sigma_{u}\right\}$ are imposed to guarantee that the derived multi-kernel $K$ is characteristic. 

The process of aligning each teacher and student is equivalent to minimizing the MMD distance between them. This can achieve intra-class cohesion of similar samples. We aggregate all such MMDs between $N$ pairs of teachers and students, and the overall alignment loss $\mathcal{L}_{align}$ in the shared MLP can be written as:
\begin{equation}
\label{align_loss}
\mathcal{L}_{align}=\sum_{i=1}^{N} \mathrm{MMD}\left(f_\mathcal{S}, f_{\mathcal{T}_{t}} \right)
\end{equation}
\subsection{Soft-target Distillation}
Apart from learning the teacher’s features, the student is also expected to produce identical or similar inferences as the teachers do. We thus also take the teachers’ predictions by feeding unlabelled input samples to them and then supervise the student’s training. As there is no annotation available for each instance $x$ in the target dataset $\mathcal{D}_s$, the predictions of pre-trained teachers can be constructed as supervision for guiding the student, named as \emph{soft-target distillation}. 
\begin{table*}[t]
\footnotesize
\small
\caption{Statistics of datasets used in this paper.}
  \centering
  \setlength{\tabcolsep}{13pt}{
    \begin{tabular}{c|c|c|c}
    \toprule
    \textbf{Dataset} & \textbf{Images} & \textbf{Categories} & \textbf{Train/Test} \\
    \midrule
    CUB-200-2011 &11,788    &200    &5,994/5,794 \\
    \midrule
    Stanford Dogs &20,580   &120    &12,000/8,580 \\
    \midrule
     Stanford Cars& 16,185 &  196  & 8,144/8,041 \\
    \midrule
    FGVC-Aircraft & 102,000 & 102   &  6,667/3,333 \\
    \bottomrule
    \end{tabular}}%
  \label{tab:datasets}%
  \vspace*{-0.25in}
\end{table*}%

Specifically, we first feed $x$ into each $T_i$ to obtain the golden label probability distribution $\Phi(x; T_i)$ in the \emph{softmax layer}, and then concatenate them together for training the student by minimizing the KL-divergence between their probability distribution:
\begin{equation}
\mathcal{L}_{std}=\sum_{x \in \mathcal{D}_s} \operatorname{KL}(\Phi(x,S) \| \Phi(x,T))
\end{equation}
where $\Phi(x,S)$ and $\Phi(x,T)$ denote the \emph{softmax} probability distribution of the student and that of the concatenated teachers for input $x$, respectively.
Considering the weighted sum of contrastive losses (including $\mathcal{L}_{intra}$, and $\mathcal{L}_{inter}$), alignment loss $\mathcal{L}_{align}$ and soft-target distillation loss $\mathcal{L}_{std}$ together, the total training objective of our CKA can be described as:
\begin{equation}
\small
\label{final_loss}
\mathcal{L}=\lambda_{intra}\mathcal{L}_{intra} + \lambda_{inter}\mathcal{L}_{inter} + \lambda_{a}\mathcal{L}_{align} + \lambda_{d}\mathcal{L}_{std}
\end{equation}
\section{Experiments}
In this section, we evaluate the proposed method on standard benchmarks and compare the results with the recent state of the arts. 
We also conduct ablation studies to validate the effect of the major components.
\subsection{Experiments Setup}
\paragraph{Datasets} We evaluate our proposed CKA on four widely used benchmarks, i.e.,  
CUB-200-2011 \cite{WahCUB_200_2011}, Stanford Cars \cite{Krause_6755945}, Stanford Dogs \cite{KhoslaYaoJayadevaprakashFeiFei_FGVC2011}, and FGVC-Aircraft \cite{maji13fine-grained}. 
The detailed statistics are summarized in Table \ref{tab:datasets}.
\paragraph{Implementation Details} 
We adopt the \emph{resnet} family \cite{HeZRS16} including \emph{resnet}-18, \emph{resnet}-34, and \emph{resnet}-50, as our model samples. 
Besides, all the teachers are first pre-trained as \cite{Krause_6755945} and fine-tuned to heterogeneous tasks. 
To construct heterogeneous tasks on the given datasets, we split all the categories into non-overlapping parts of equal size to train the teachers. 
The trained teacher model weights are frozen during the student training process. 
In student training phrase, data augmentation is performed via Random ResizedCrop, Random ColorJitter, Random HorizontalFlip, and Random GaussianBlur while in testing, Center Crop is used.
During training, the learning rate is set to 0.0005, and the cosine decay is used; the weight decay is set to 0.0005; Adam is used as the optimizer, and the batch size is set to 64; a total of 100 epochs are trained. 
All experiments are completed with GPUs of RTX 2080 Ti 11GB and CPUs of Intel. 
There are several hyper-parameters involved in our method, including $\alpha$ in Eqn. \ref{margin_contrast}, set to 0.4, for alleviating the dominance of negative sample pairs;  $\lambda_{intra}$, $\lambda_{inter}$, $\lambda_{a}$ and $\lambda_{d}$ for the final CKA loss in Eqn. \ref{final_loss}, are set to $\lambda_{intra}$ = $\lambda_{inter}$ = $\lambda_{d}$ = 1 and $\lambda_{a}$ = 10. 
\begin{table*}[t]
  \centering
  \small
  \caption{Comparison of different methods on comprehensive classification tasks. 
  Best results are shown in bold.}
\label{tab:benchmarks_comparison}
    \begin{tabular}{l|c|cccc|c}
    \toprule
    \textbf{Method} & \textbf{Size} & \textbf{Dogs} & \textbf{Cars} & \textbf{CUB} & \textbf{Aircraft} & \textbf{Average}\\
    \midrule
    Supervised & 163M & 83.62 $\pm$ 0.00 & 89.64 $\pm$ 0.00  &  72.68 $\pm$ 0.00   & 82.78 $\pm$ 0.00  & 82.14 \\
    \midrule
    Teacher1 & 130M & 66.64 $\pm$ 0.00 & 70.33 $\pm$ 0.00    &  65.37 $\pm$ 0.00  & 63.01 $\pm$ 0.00 & 66.80\\
    Teacher2 & 240M    & 72.03 $\pm$ 0.00 & 87.85 $\pm$ 0.00    &  66.12 $\pm$ 0.00     & 81.12 $\pm$ 0.00  & 76.60\\
    Ensemble & 370M    & 73.90 $\pm$ 0.22 & 77.08 $\pm$ 0.64 & 68.25 $\pm$ 0.00  & 75.76 $\pm$ 0.00  & 73.38 \\
    \midrule
    Vanilla KD & 240M  & 76.16 $\pm$ 0.60 & 80.39 $\pm$ 0.31 & 69.94 $\pm$ 0.79  & 78.00 $\pm$ 0.01 &76.06 \\
    CFL   & 240M     & 76.23 $\pm$ 0.26 & 81.12 $\pm$ 0.21 & 70.67 $\pm$ 0.97  & 79.98 $\pm$ 0.22 &76.86\\
    \midrule
    CKA-Intra & 240M     &78.89 $\pm$ 0.59    &  82.33 $\pm$ 0.31     & 71.07 $\pm$ 0.04     &  79.02 $\pm$ 0.21 &77.71\\
     CKA-Inter & 240M     &79.72 $\pm$ 0.60    &  \textbf{82.95 $\pm$ 1.20}     & \textbf{71.49 $\pm$ 0.25}     &  80.45 $\pm$ 0.51 & 78.46\\
      CKA & 240M     &\textbf{79.76 $\pm$ 0.09}    &  82.88 $\pm$ 0.21     & 71.32 $\pm$ 0.55\     &  \textbf{80.78 $\pm$ 0.08} & 78.45\\
    \bottomrule
    \end{tabular}%
    \vspace*{-0.25in}
\end{table*}%
\paragraph{Compared Methods}
We implement various baselines to evaluate the effectiveness of our proposal, which are categorized as:
(1) \emph{Original Teacher}: The teacher models are used independently for prediction. We set the probabilities of classes out of the teacher specialty to zeros. 
(2) \emph{Ensemble}: The output logits of teachers are directly concatenated for predictions over the union label set. 
(3) Vanilla KD \cite{HintonVD15}: The student is trained to mimic the soft targets produced by logits combination of all teacher models, via minimizing the vanilla KL-divergence objective. 
(4) CFL \cite{LuoWFHTS19}: CFL first maps the hidden representations of the student and the teachers into a common feature space. 
The student is trained by aligning the mapped features to that of the teachers, with supplemental supervision from the logits combination. 
We also include a supervised learning method, which trains the student with labeled data for a better understanding of the performance. We compare the average accuracy of each method in three random experiments.
\subsection{Quantitative Analysis}
We compare our proposed method CKA with SOTA on above-mentioned classification datasets. The experiment results and corresponding model size are listed in Table \ref{tab:benchmarks_comparison}. 
Our findings are: (1) Simple baselines can be seriously affected by incomplete datasets and annotations, showing that it is necessary to conduct amalgamation. 
(2) CFL cannot achieves consistent improvements on comprehensive tasks, demonstrating the instability of supervision based on simple feature alignments. 
(3) Our proposed CKA and its variants outperform the previous baseline models on all the datasets, and the average accuracy of CKA-Inter is achieves a 1.60 points gain over the best performing baseline model. 
On the FGVC-Aircraft dataset, the knowledge consolidation accuracy of CKA reached 80.78\% without label information, approaching that of supervised learning methods.
We attribute this success to the fact that CKA provides the student with natural semantic relevance estimated on the sample set based on contrastive losses, and the intra-class cohesion and inter-class separation methods effectively transfer feature-level knowledge.
Furthermore, supervisory contradictions from incomplete teachers are avoided by soft labels at task-level amalgamation.
These promising results indicate that our CKA framework produces better supervisions for training the student model, yields great potentials for model reusing.
\subsection{Ablation Study}
We conduct ablation studies to investigate the contribution of the contrastive losses and soft-target distillation loss described in our proposed approach.


 For margin-based intra-model contrastive loss, we compare the performances by turning them on and off. 
 For inter-model loss between teacher-student pairs, on the other hand, we define three different distances in Eqn. \ref{eq:distance}, including euclidean distance, cosine and MMD distance. 
 We summarize the comparative results in Table \ref{tab:ablation_study}, where we observe that the CKA-Inter with MMD distance yields better performance than others. 
Moreover, CKA and its variants also improve with a large room over KD and CFL, validating the complement of contrastive losses and flexibility of soft-target loss.
\begin{table}[t]
\caption{Ablation analysis of CKA. Remove modules lead to deteriorated performance}
\label{tab:ablation_study}
\footnotesize
\begin{tabular}{@{}ccccccccc@{}}
\toprule
\multirow{2}{*}{\textbf{Method}} &
  \multirow{2}{*}{\begin{tabular}[c]{@{}c@{}}Vanilla \\ KD$_{kd}$\end{tabular}} &
  \multirow{2}{*}{CFL} &
  \multirow{2}{*}{CKA} &
  \multirow{2}{*}{\begin{tabular}[c]{@{}c@{}}W/O Inter-model \\ loss\end{tabular}} &
  \multicolumn{3}{c}{W Inter-model loss} &
  \multirow{2}{*}{\begin{tabular}[c]{@{}c@{}}W/O Intra-model \\ loss\end{tabular}} \\ \cmidrule(lr){6-8}
                 &       &       &       &       & Euclidean & Cosine & MMD   &       \\ \midrule
\textbf{Cars}     & 80.22 & 81.12 & 82.88 & 80.04 & 82.33     & 82.95  & 83.21 & 82.33 \\ \midrule
\textbf{Aircraft} & 78.00 & 79.98 & 80.78 & 77.97 & 80.21     & 80.45  & 81.42 & 79.02 \\ \bottomrule
\end{tabular}
\vspace*{-0.25in}
\end{table}

\begin{table}[b]
\vspace*{-0.25in}
\caption{Result of merging heterogeneous teachers with different architectures is demonstrated by Stanford Dogs dataset. 
}
\small
\label{tab:cross_architecture_comparison}
\begin{tabular}{@{}cc|ccc|ccc@{}}
\toprule
\multicolumn{2}{c|}{\textbf{Teachers}} &
  \multicolumn{1}{c}{$\mathcal{T}_1$: \emph{restnet-18}} &
  \multicolumn{2}{c|}{$\mathcal{T}_2$: \emph{restnet-34}} &
  \multicolumn{2}{c}{$\mathcal{T}_1$: \emph{restnet-50}} &
  \multicolumn{1}{c}{$\mathcal{T}_2$: \emph{restnet-34}} \\ \midrule
\multicolumn{2}{c|}{\textbf{Method}}     & Vanilla KD  & CFL   & CKA   & Vanilla KD & CFL   & CKA   \\ \midrule
\multicolumn{1}{c|}{\multirow{2}{*}{\textbf{\begin{tabular}[c]{@{}c@{}}Student\\ Net\end{tabular}}}} &
  \emph{resnet-34} &
  80.67 &
  81.09 &
  82.54 &
  80.54 &
  81.23 &
  82.08 \\
\multicolumn{1}{c|}{} & \emph{resnet-50} & 82.04       & 82.25 & 83.18 & 82.62     & 84.55 & 85.21 \\ \bottomrule
\end{tabular}
\vspace*{-0.3in}
\end{table}
\subsection{Results in Challenging Settings}
\subsubsection{CKA with Heterogeneous Teachers}
We further consider merging knowledge from heterogeneous teachers with different structures. 
Specifically, we random select two different \emph{resnet} architectures as the teachers, respectively. 
The results are listed in Table \ref{tab:cross_architecture_comparison}. 
We find that while a larger student tends to perform better, indicating that the wider and larger the model, the more complete the knowledge can be learned. 
Our CKA achieves the best results on the Stanford Dogs, showing its effectiveness for heterogeneous teachers.
\begin{table}[t]
  \centering 
  \footnotesize
   \caption{ Results of merging from teacher models with different knowledge domains and in a cross-dataset scenario of Stanford Cars and FGVC-Aircraft.}
    \small
\label{tab:cross_data_comparison}
\setlength{\tabcolsep}{10pt}{
    \begin{tabular}{l|cc|c}
    \toprule
    \textbf{Method} &
    \textbf{$\mathcal{T}_1$: Stanford Cars} & \textbf{$\mathcal{T}_2$: FGVC-Aircraft} & \textbf{Merge}\\
    \midrule
    Supervised & 89.64 $\pm$ 0.00 & 82.78 $\pm$ 0.00 &  86.90 $\pm$ 0.00\\
    \midrule
    Teacher1 & 89.64 $\pm$ 0.00 & —   &  — \\
    Teacher2 & — & 78.00 $\pm$ 0.00   &  — \\
    Ensemble & — & — & 82.08 $\pm$ 0.54 \\
    \midrule
    Vanilla KD & 85.26 $\pm$ 0.25 & 80.31 $\pm$ 0.85 & 83.76 $\pm$ 0.59 \\
    CFL & 87.99 $\pm$ 0.73 & 84.22 $\pm$ 0.48 & 86.76 $\pm$ 0.46 \\
    \midrule
    CKA-Intra &88.95 $\pm$ 0.00    &84.93 $\pm$ 0.58     & 87.75 $\pm$ 0.60\\
     CKA-Inter &\textbf{89.48 $\pm$ 0.31}    &84.91 $\pm$ 0.59     & 88.11 $\pm$ 0.79\\
      CKA &89.28 $\pm$ 0.75    &\textbf{85.78 $\pm$ 0.07}     & \textbf{88.21 $\pm$ 0.50}\\
    \bottomrule
    \end{tabular}}%
%
\end{table}%
\subsubsection{CKA with Heterogeneous Teachers for Cross-Dataset}
Specifically, we pretrain distinct-task teacher models on different datasets separately and then train a student to perform classification over the union label set of both datasets. 
The results of merging knowledge from two combined datasets, Stanford Cars and FGVC-Aircraft are listed in Table \ref{tab:cross_data_comparison}. 
\emph{resnet-34} is adopted for training student in the cross-dataset setting. 
Our CKA still outperforms previous baseline models in this settings. Interestingly, we find that the performance of CKA is superior to all baselines and even to the results of supervision. 
We speculate that the reason is that the correlation between classes in different datasets is weak and the data classification categories are complex, which is prone to confusion by label supervision alone. 
In contrast, our CKA uses contrast loss to compute the distance between samples, which is more robust and discriminative.

\section{Conclusion}
In this paper, we explore knowledge amalgamation for unsupervised classification tasks for promoting better model reuse. We present a principled framework CKA, in which contrastive losses and alignment loss are designed to enlarge the distance between feature representations of samples from different categories and decrease that of samples from the same categories, as a self-supervised way to guide the student to learn discriminative features. Besides, we present a soft-target distillation loss to efficiently and flexibly transfer the dark knowledge in the task-level amalgamation. Experiments on several benchmarks demonstrate our CKA can substantially outperform strong baselines. More extensive investigations show that CKA is generalizable for challenging settings, including merging knowledge from heterogeneous teachers, or even cross-dataset teachers.

\bibliographystyle{icann23.bst}
\bibliography{icann23.bib}
\end{document}